\RequirePackage{luatex85}

\documentclass[letterpaper, 10pt, conference]{ieeeconf}  % Comment this line out if you need a4paper

\IEEEoverridecommandlockouts    % This command is only needed if you want to use the \thanks command

\usepackage{graphics} % for pdf, bitmapped graphics files
\usepackage{epsfig} % for postscript graphics files
\usepackage{mathptmx} % assumes new font selection scheme installed
\usepackage{times} % assumes new font selection scheme installed
\usepackage{amsmath} % assumes amsmath package installed
\usepackage{amssymb}  % assumes amsmath package installed

\usepackage{url} % To have nice URL
\usepackage{subcaption} % For subfigures
\usepackage[font=small,skip=1pt]{caption}

\usepackage{tikz}  % for diagrams
\usetikzlibrary{arrows.meta}
\usepackage{pgfplots} % for nice plots
\usepgfplotslibrary{groupplots}
\usetikzlibrary{bayesnet} % to draw bayesnet
\usepackage{standalone} % to include .tex file

\usepackage{adjustbox}
\usepackage{booktabs} % table formatting
\usepackage{tabularx}
\usepackage{etoolbox}
\robustify\bfseries

\pgfplotsset{compat=newest}
\pgfplotsset{every axis legend/.append style={%
		style={at={(0,1)},anchor=north west},
		cells={anchor=west}},
}

\usepackage{siunitx} % to handle SI units
\usepackage[capitalize]{cleveref} % Internal references

\usepackage{booktabs}

\usepackage[backend=bibtex,style=ieee,sortcites]{biblatex}
\AtEveryBibitem{
	\clearfield{doi}
	\clearlist{address}
	\clearname{editor}
	\clearlist{organization}
	\clearfield{url}  
	\clearfield{doi}  
	\clearlist{location}
	\clearlist{issn}
	\clearlist{ibn}
	\ifentrytype{inproceedings}{
		\clearfield{pages}
		\clearfield{publisher}
	}{}	
}

\makeatletter
\def\blx@maxline{77}
\makeatother

\renewbibmacro{finentry}{%
	\iffieldequalstr{entrykey}{idm}%<- key after which you want the break
	{\finentry\newpage}
	{\finentry}}

\AtBeginBibliography{\small}

\title{\LARGE \bf
Belief State Planning for Autonomously Navigating Urban Intersections
}

\author{Maxime Bouton$^{1}$, Akansel Cosgun$^{2}$, and Mykel J. Kochenderfer$^{1}$   % <-this % stops a space
\thanks{*This work was supported by the Honda Research Institute.}%
\thanks{$^{1}$ Maxime Bouton and Mykel J. Kochenderfer are with the Department of Aeronautics and Astronautics, Stanford University, Stanford CA 94305, USA,
        {\tt \{boutonm,mykel\}@stanford.edu}.}%
\thanks{$^{2}$ Akansel Cosgun is with the Honda Research Institute, 375 Ravendale Dr., Mountain View, CA 94043, USA, 
        {\tt acosgun@hra.com}.}%
}

\begin{document}

%%% MUST BE UP TO 6 PAGES %%%

\maketitle
\thispagestyle{plain}
\pagestyle{plain}

%%%%%%%%%%%%%%%%%%%%%%%%%%%%%%%%%%%%%%%%%%%%%%%%%%%%%%%%%%%%%%%%%%%%%%%%%%%%%%%%
\begin{abstract}

Urban intersections represent a complex environment for autonomous vehicles with many sources of uncertainty. The vehicle must plan in a stochastic environment with potentially rapid changes in driver behavior. Providing an efficient strategy to navigate through urban intersections is a difficult task. This paper frames the problem of navigating unsignalized intersections as a partially observable Markov decision process (POMDP) and solves it using a Monte Carlo sampling method. Empirical results in simulation show that the resulting policy outperforms a threshold-based heuristic strategy on several relevant metrics that measure both safety and efficiency.

\end{abstract}

%%%%%%%%%%%%%%%%%%%%%%%%%%%%%%%%%%%%%%%%%%%%%%%%%%%%%%%%%%%%%%%%%%%%%%%%%%%%%%%%
\section{Introduction}

Intersections account for 40\% of driving accidents and represent a major challenge for automated driving \cite{Statistics2008}. Handling intersections involves planning under uncertainty with respect to driver behavior. One must be able to infer the goals of the other agents and anticipate rapid maneuver changes.  
Another difficulty is ensuring the satisfaction of multiple conflicting objectives including the risk of accident, the time to cross the intersection, the comfort of the passengers, and the disturbance caused to other drivers. The relative importance of the objectives varies by driver. %Some drivers minimize the time to cross in expense of a greater risk of collisions while potentially causing traffic disturbances.
The variety of users present in urban environments as well as complicated traffic rules also make intersections difficult to handle. 

Several approaches have been employed to address intersection crossing.
One approach involves hand-engineering hierarchical state machines that attempt providing explicit strategies for all possible situations. 
These state machines are useful for solving simple driving problems, but rely heavily on the designer to anticipate how to best handle different situations in advance. Hierarchical state machines were used in the DARPA urban challenge \cite{darpa2} and almost lead to an accident at an intersection \cite{darpa1}. 

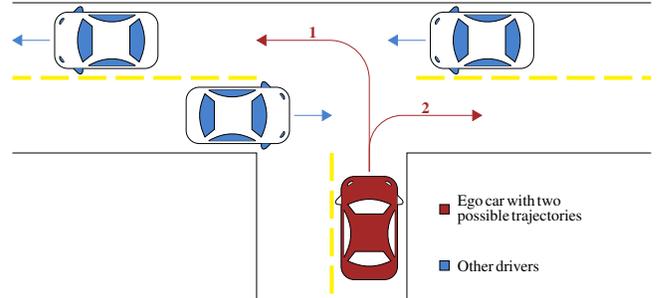
\begin{figure}
	\centering
	\resizebox{\columnwidth}{!}{
	\begin{tikzpicture}

\definecolor{myblue}{RGB}{72,130,206}
\definecolor{myred}{RGB}{165,40,40}

	%upper border
	\draw [very thick] (-5,16) -- (29,16);
	
	%lower left 
	\draw [very thick] (-5,8) -- (8,8) -- (8,0);
	
	%lower right  
	\draw [very thick] (16,0) -- (16,8) -- (29,8);
	
	%yellow lines 
	\draw [line width=6, color=yellow] (-5,12) -- (-4,12);
	\draw [line width=6, color=yellow] (-3.5,12) -- (-2,12);
	\draw [line width=6, color=yellow] (-1.5,12) -- (0,12);
	\draw [line width=6, color=yellow] (0.5,12) -- (2,12);
	\draw [line width=6, color=yellow] (2.5,12) -- (4,12);
	\draw [line width=6, color=yellow] (4.5,12) -- (6,12);
	\draw [line width=6, color=yellow] (6.5,12) -- (8,12);
	
	\draw [line width=6, color=yellow] (16.5,12) -- (18,12);
	\draw [line width=6, color=yellow] (18.5,12) -- (20,12);
	\draw [line width=6, color=yellow] (20.5,12) -- (22,12);
	\draw [line width=6, color=yellow] (22.5,12) -- (24,12);
	\draw [line width=6, color=yellow] (24.5,12) -- (26,12);
	\draw [line width=6, color=yellow] (26.5,12) -- (28,12);
	\draw [line width=6, color=yellow] (28.5,12) -- (29,12);	
	
	\draw [line width=6, color=yellow] (12,0.5) -- (12,2);
	\draw [line width=6, color=yellow] (12,2.5) -- (12,4);
	\draw [line width=6, color=yellow] (12,4.5) -- (12,6);
	\draw [line width=6, color=yellow] (12,6.5) -- (12,8);

	%insert ego car 
	\node [inner sep=0pt] (egoCar) at (14,4)
	{\includestandalone[]{red_car}};
	
	%insert human car 
	\node [inner sep=0pt] (hcar1) at (0,14)
	{\includestandalone[angle=90]{blue_car}}; 
	\node [inner sep=0pt] (hcar2) at (7,10)
	{\includestandalone[angle=-90]{blue_car}}; 
	
	\node [inner sep=0pt] (hcar3) at (20,14)
	{\includestandalone[angle=90]{blue_car}};
	
	%left trajectory
	\draw [thick, color=myred, rounded corners=60,->,>=Triangle,-{>[scale=3.0]}] (14,7) -- (14,14) -- node[above]{\Huge \textbf{1}} (8,14);
	
	%right trajectory
	\definecolor{myred}{RGB}{165,40,40}
	\draw [thick, color=myred, rounded corners=50,->,>=Triangle,-{>[scale=3.0]}] (14,7) -- (14,10) -- node[above]{\Huge \textbf{2}} (20,10);
	
	%speed of the other drivers
	\draw [thick, color=myblue,->,>=Triangle,-{>[scale=3.0]}] (-3,14) -- (-5,14);
	\draw [thick, color=myblue,->,>=Triangle,-{>[scale=3.0]}] (17,14) -- (15,14);
	\draw [thick, color=myblue,->,>=Triangle,-{>[scale=3.0]}] (10,10) -- (12,10);

	%labels
	\node (egoLab) [draw,minimum width=0.5cm,minimum height=0.5cm,fill=myred] at (18,5) {};
	\node (egoLab2) [text width=7cm] at (22.2,5) {\Huge Ego car with two \\ possible trajectories};
	
	\definecolor{myblue}{RGB}{72,130,206}
	\node (humLab) [draw,minimum width=0.5cm,minimum height=0.5cm,fill=myblue] at (18,2) {};
	\node (humLab2) [text width=10cm] at (23.7,2) {\Huge Other drivers};

\end{tikzpicture}
	}
	\caption{The objective of the autonomous vehicle is to decide on the acceleration to apply along a given path. Two different scenarios are considered: right-turn and left-turn}
	\label{fig:scenario}
\end{figure}

% describe challenges
Learning-based methods can help reduce the burden on the designer for developing robust decision strategies. One type of learning approach known as behavioral cloning involves learning a policy from a human driver \cite{deep1,deep2}. Some behavioral cloning approaches attempt to directly map sensor readings (e.g., raw pixels from a camera \cite{deep1}) to driving commands. These approaches rely on a large amount of data and are unlikely to perform better than the human driver used for training. %The robustness of these approaches still needs to be proven \cite{deep1}.

% litt review
Another category for developing intersection crossing strategies involves planning with respect to a mathematical model of the problem. A partially observable Markov decision process (POMDP) is a standard model for sequential problems with stochastic state transitions and sensor uncertainty \cite{pomdp1,pomdp2,pomdp3}. 
One of the challenges in this approach is in representing and modeling the problem in a way that allows the planning algorithm to be tractable  \cite{pomdp4}. Offline planners compute an approximately optimal strategy over the entire state space, prior to execution. Online planners, on the other hand, compute the best action to take at the current time step. A popular online algorithm is partially observable Monte Carlo Planning (POMCP), which relies on sampling from a generative model \cite{pomcp}. For the intersection problem in this paper, we augment POMCP with progressive widening \cite{dpw} to accommodate the continuous state space.

% assumptions and what we are going to do
The objective of this work is to develop an online decision making algorithm to cross an urban intersection autonomously. By modeling the problem as a POMDP, the autonomous system can dynamically change its decision to adapt to the behavior of other agents. As shown in \cref{fig:scenario}, vehicles are at an unsignalized T-junction with traffic flowing in both directions. The autonomous vehicle is initially stopped at the intersection and tries to turn left or right. The nominal path is assumed to be generated by a high-level task planner and the proposed planner computes the acceleration profile along this path. Although we consider noisy position and velocity measurements, we do not consider sensor limitations such as occlusions. Finally, the behavior of the vehicles are represented by internal states that are not directly observable, but rather they are estimated using an interacting multiple model (IMM) filter \cite{imm}.

%%%%%%%%%%%%%%%%%%%%%%%%%%%%%%%%%%%%%%%%%%%%%%%%%%%%%%%%%%%%%%%%%%%%%%%%%%%%%%%%
\section{Proposed Approach}

\subsection{POMDP Background}

A partially observable Markov decision process (POMDP) is a mathematical framework for sequential decision making under uncertainty. It is formally characterized by a tuple $(\mathcal{S},\mathcal{A},\mathcal{O},T,O,R,\gamma)$, where $\mathcal{S}$ is the state space, $\mathcal{A}$ the action space, $\mathcal{O}$ is the observation space, $T$ is a transition model, $O$ is an observation model, $R$ a reward model, and $\gamma$ is a discount factor \cite{dmu}.

Uncertainty is represented by the transition model and the observation model. An important source of uncertainty in the context of autonomous driving is the behavior of the other drivers \cite{pomdp3,pomdp4}. From a state $s\in \mathcal{S}$ of the environment, the agent takes an action $a\in \mathcal{A}$ to maximize the expected accumulation of reward $r(s,a)$ over time. The state $s$ will then transition to $s'$ with probability $T(s',s,a) = \Pr(s'\mid s,a)$ under the Markov assumption that the state $s'$ only depends on the previous state.

The agent has uncertain knowledge about the state of the environment  and maintains a belief state $b\in \mathcal{B}$. The belief state is a probability distribution over all possible states, $b: \mathcal{S}\mapsto[0,1]$, and $b(s)$ represents the probability of being in state $s$.
Belief state planning involves finding a policy that maps belief states to actions in a way that maximizes the expected discounted accumulation of reward over time. 

%$\pi: \mathcal{B} \mapsto \mathcal{A}$ that maximizes the expected accumulated reward  $U(b)$:
%\begin{align}
%U^\pi &= \mathbf{E}[\sum\limits_{t=0}^{\infty} \gamma^tr(s_t,\pi(b_t))]\\
%\pi^* &= \mathrm{arg}\max_{\pi} U^\pi
%\end{align}
%In the equation above, the state and belief are indexed by the time step $t$.

\subsection{Intersection Navigation Problem}\label{sec:model}

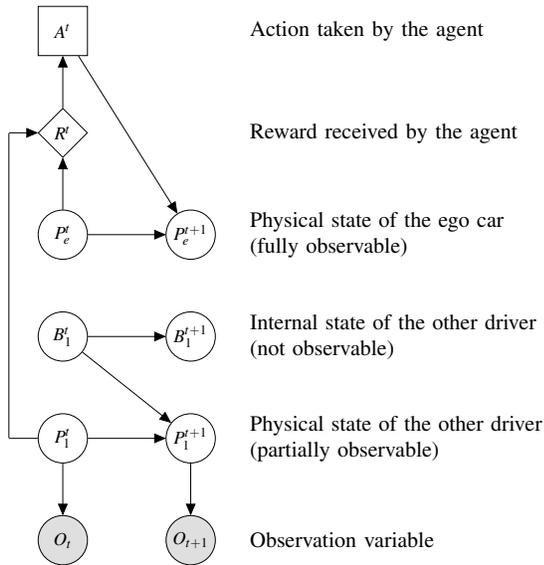
\begin{figure}
	\vspace{5pt}
    \centering
    \resizebox{0.9\columnwidth}{!}{
    % nodes for drawing
\tikzstyle{fake} = [rectangle, fill=black,minimum size=0.pt,inner sep=0pt]

\begin{tikzpicture}
% Nodes at time t
\node[rectangle,draw=black, minimum size=27pt](at){$A^t$};
\node[det,minimum size=27pt,below=of at](rt){$R^t$};
\node[latent,minimum size=27pt, below=of rt](pet){$P_e^t$};
\node[latent,minimum size=27pt, below=of pet](bt){$B_1^t$};
\node[latent,minimum size=27pt, below=of bt](pht){$P_1^t$};
\node[obs,minimum size=27pt, below=of pht](ot){$O_t$};

% Nodes at time t+1
\node[latent,minimum size=27pt, right=1.5 of pet](pet1){$P_e^{t+1}$};
\node[latent,minimum size=27pt, right=1.5 of bt](bt1){$B_1^{t+1}$};
\node[latent,minimum size=27pt, right=1.5 of pht](pht1){$P_1^{t+1}$};
\node[obs,minimum size=27pt, right=1.5 of ot](ot1){$O_{t+1}$};

%Nodes for drawing
\node[fake, node distance=0., left=0.55 of pht](a){};
\node[fake, node distance=0., left=0.55 of rt](b){};

%\node[rectangle, fill=black,minimum size=0pt, inner
%sep=0pt, node distance=0., left=0.7 of ot](c){};
%\node[rectangle, fill=black,minimum size=0pt, inner
%sep=0pt, node distance=0., left=0.7 of ot1](d){};

%labels 
\node[right=3. of at, font=\large](alab){Action taken by the agent};
\node[right=3. of rt, font=\large](rlab){Reward received by the agent};
\node[right=3. of pet, font=\large, text width=6cm](pelab){Physical state of the ego car\\(fully observable)};

\node[right=3. of bt, font=\large,text width=6cm](blab){Internal state of the other driver\\(not observable)};

\node[right=3. of pht, font=\large, text width=6cm](phlab){Physical state of the other driver\\(partially observable)};

\node[right=3. of ot, font=\large](olab){Observation variable};

%edges
\edge{rt}{at};
\edge{pet}{rt};
\edge[-]{pht}{a};
\edge[-]{a}{b};
\edge{b}{rt};

%\edge[-]{ot}{c}
\edge{pht}{ot}
\edge{pht1}{ot1};
\edge{at}{pet1};
\edge{pet}{pet1};
%%\edge{pet}{pht1};
\edge{pht}{pht1};
%%\edge{pet1}{pht1};
\edge{bt}{pht1};
\edge{bt}{bt1};
%
%\edge[-]{ot1}{d}
%\edge{d}{pht1}
%\edge{ot1}{bt1};

\end{tikzpicture}
    }
    \caption{Structure of the problem represented as a Bayesian network, the three components of the state variable are assumed independent.}
    \label{fig:bayesnet}
\end{figure}

The structure of the problem can be represented by the Bayesian network in \cref{fig:bayesnet}. The round nodes represent state variables that change over time. The diamond-shaped node corresponds to the reward received and the square node represents the action taken at a given time step. 

\subsubsection{State Space}

The state in the POMDP includes both the physical state of the environment and the behavior of other drivers. The intersection geometry is assumed to be known to the planner.
Three sets of variables are used to define the state space:
\begin{itemize}
\item $P_e^t = (x_e^t,y_e^t,\theta_e^t,v_e^t,a_e^t)$ describes the physical state of the ego car at time $t$.
\item $P_i^t = (x_i^t,y_i^t,\theta_i^t,v_i^t,a_i^t),\; i=1,\ldots,n$ describes the physical state of the $n$ other cars in the intersection at time $t$.
\item $B_i^t$ describes the internal state of the other drivers, characterizing their behavior at time $t$. In the experiments used in this paper, the internal state may correspond to one of two different models.
\end{itemize}
Here, $x$ and $y$ represent the position in a Cartesian frame, $\theta$ represents the vehicle orientation, $v$ is the speed, and $a$ is the magnitude of the acceleration. Variables with subscript $e$ are associated with the ego car, and variables with subscript $i$ correspond to the $i$th vehicle. The relationships between the variables are illustrated by \cref{fig:bayesnet}. To simplify the structure of the problem we assume independence between the ego car physical state and the other vehicles' physical states. 

\subsubsection{Action space}
The objective of the planner is to compute the acceleration profile along the desired path (left-turn or right-turn as illustrated in \cref{fig:scenario}). Strategic maneuvers such as hard braking, moderate braking, maintaining constant speed and accelerating can be represented by a finite set of acceleration and deceleration action: $\{\SI{-4}{\meter\per\second\squared},\SI{-2}{\meter\per\second\squared},\SI{0}{\meter\per\second\squared},\SI{2}{\meter\per\second\squared}\}$.

\subsubsection{Process Model}

The input to the problem is the ego path; either a left or right turn. The motion of the ego car along this path is only controlled by an acceleration input. Given the shape of the trajectory, it is more convenient to use polar coordinates. The kinematic equations used to update the ego car state are as follows: 
\begin{equation}
\left\{
\begin{array}{l}
x_e^{t+1} = x_e^t + v_e^t\sin(\theta_e^t)\delta t + a_e^t\sin(\theta_e^t)\frac{\delta t^2}{2} \\
y_e^{t+1} = y_e^t + v_e^t\cos(\theta_e^t)\delta t + a_e^t\cos(\theta_e^t)\frac{\delta t^2}{2} \\
v^{t+1} = v_e^t + a_e^t\delta t
\end{array}
\right.
\label{eq:process}
\end{equation}
The orientation $\theta$ is updated according to the desired trajectory. In the equations, $\delta t$ is the time step between decisions.

Two different kinematic models were used to model the behavior of the other drivers: constant velocity (CV) and constant acceleration (CA) \cite{imm}. These models can describe various behavior including braking (CA model with negative acceleration) or maintaining speed (CV model). The state transition function follows linear Gaussian dynamics:
\begin{equation}
\Pr(P_i^{t+1} \mid P_i^t) = \mathcal{N}(P_i^{t+1} \mid \mathbf{T}P_i^t,\mathbf{Q})
\end{equation}
where $\mathbf{T}$ is the state transition matrix and $\mathbf{Q}$ the process noise. These matrices are different for each kinematic model and follow the equations of \citeauthor{imm} \cite{imm}. Process noise matrices are characterized by a spectral density $\sigma_{\{CV,CA\}}$. The Gaussian dynamics provide a suitable representation of the world as it is describing continuous variables with a minimal amount of information (mean and covariance) so that the problem remains computationally feasible. 

The internal state $B_i^t$ can have one of two values corresponding to the two possible kinematic models. Given the value of $B_i^t$, the variable $P_i^t$ is updated using \cref{eq:process}. At each time step, we assume that the behavior can change according to a switching probability matrix $p$ where $p_{ij}$ is the probability from switching to model $j$ from model $i$. No prior knowledge of the path of the other drivers is assumed.

\subsubsection{Observation Model}\label{sec:observation-model}

The observation space describes what we can measure in the environment. The vehicle is assumed to have a perfect knowledge of its physical state. We assumed that position and velocity of other drivers are partially observable while their acceleration and their behavior are internal states that cannot be measured. Finally, the orientation of the other cars is assumed to be known. The observation space is defined as follows:
\begin{equation}
O_i^t = (z^t_{x_i},z^t_{y_i},\theta_i^t,z^t_{v_i}),\quad i=1,\dots, n
\end{equation}
where the components are the measured position, the orientation, and the velocity of the $i$th vehicle at time $t$.

To have a simple representation of the observation distribution, we model the sensor measurements using a Gaussian distribution:
\begin{equation}
\Pr(O_i^t\mid P_i^t) = \mathcal{N}(O_i^t \mid \mathbf{H} P_i^t,\mathbf{R})
\label{eq:obs-dis}
\end{equation}
where $\mathbf{H}$ is the observation model matrix and $\mathbf{R}$ is the observation noise matrix. We assume that the measurements are independent ($\mathbf{R}$ diagonal) and characterized by the standard deviations $\sigma_p$ for the position measurement and $\sigma_v$ for the velocity measurement.

\subsubsection{State Estimation}

Since the state is factored into three independent variables, the belief is defined in a similar way:
\begin{itemize}
\item The ego car physical state, assumed perfectly known.
\item A distribution over the vehicle physical state. From the measurements, we can maintain an estimate of the other vehicles' state. We assume that the physical states follow a Gaussian distribution $\mathcal{N}(\hat{P_i^t},\hat{\Sigma_i^t})$, where $\hat{P_i^t}$ and $\hat{\Sigma_i^t}$ are the estimated mean and covariance of the physical state of vehicle $i$ at time $t$.
\item A distribution over the two possible kinematic models $\{\mu_{1i}^t,\mu_{2i}^t\}$ representing the probability of car $i$ following the CV model and the probability of car $i$ following the CA model, respectively. 
\end{itemize}

To infer which of the following models the cars are following, we used an Interacting Multiple Model (IMM). IMMs have been used in tracking applications and pedestrian intention prediction \cite{ped,mt}, as well as for lane changing detection in autonomous driving \cite{adimm}.  The IMM mixes two Kalman filters for both kinematic models CV and CA and update both the state estimation and the model probability distribution at each time step given an observation.

The IMM algorithm consists of three steps: mixing, filtering and combining. The first step computes an estimate of the state with respect to the two transition models and from these estimates computes two mixed inputs (a linear combination of both). The two mixed inputs are then filtered using a classic Kalman filter \cite{imm}. 

In the POMDP context, we used the IMM as the belief updater. It takes as input a belief state and an observation and returns the updated belief state. It acts only on the partially observable part of the belief space, i.e. not on the ego car state. 

One of the subtleties of the problem is that we cannot directly measure the intentions of other drivers. Since we assume that the drivers are following one among two possible Gaussian dynamics, the IMM is particularly well suited to estimate the states.

\subsubsection{Reward Model}

The agent is rewarded for reaching a final position in the intersection and receives a small penalty for each action and a large penalty for collision.

%%%%%%%%%%%%%%%%%%%%%%%%%%%%%%%%%%%%%%%%%%%%%%%%%%%%%%%%%%%%%%%%%%%%%%%%%%%%%%%%
\section{Online Belief State Planning}

Methods for computing optimal policies for a POMDP can be divided into two categories: offline and online \cite{dmu}. Offline methods compute the policy over the entire state space prior to execution in the environment. Hence, they typically do not scale to high dimensional problems. In our problem formulation, we have five continuous variables for each vehicle in the intersection. Computing the policy over the entire belief state space is intractable. Moreover, it is likely that many states will never be encountered when interacting  in the environment.

Online methods plan from the current belief state up to a certain horizon. As a consequence, online planning algorithms consider only the states reachable from the current belief and at each time step the solver computes the (approximately) optimal action. The best action is typically recomputed after each interaction with the environment.

\subsection{POMCP}

Since the state space is continuous, we use a sampling-based method known as 
Partially Observable Monte Carlo Planning (POMCP). POMCP is an extension of the Upper Confidence Tree (UCT) algorithm with partially observable state variables. The algorithm takes as input a belief state. From this belief state, it will build a tree where each node represents a history $h$, which is a sequence of actions and observations. Each node is sampled using the model described in \cref{sec:model}. The construction of the tree involves iterating through the following three steps many times:
\begin{itemize}
\item Expansion: If the node is not in the tree, we explore the outcome of all the possible actions and initialize $N(h,a)$ and $Q(h,a)$, which are the number of times we visited the node $h$ taking action $a$ and the associated value function.
\item Rollout: We simulate up to a desired depth according to a rollout policy.
\item Search: If the sampled state is already in the tree, we choose the action that maximizes $Q(h,a)+c\sqrt{\frac{N(h)}{N(h,a)}}$, where $N(h)$ is the number of times the history
was visited and $N(h, a)$ is the number of times the action and history was visited. The parameter $c$ controls the balance between exploration and exploitation.
\end{itemize}
After each iteration, the information is then propagated up to the root node. The POMCP algorithm converges to the optimal policy as the number of tree queries increases.

\subsection{Planning in Continuous State Space}

One of the drawbacks of POMCP is that it cannot handle continuous state spaces. When sampling a continuous variable from the initial belief, the probability of visiting the same state twice is infinitesimally small, resulting in a very wide tree with a depth of one. One way to address this issue is to use progressive widening \cite{dpw}.

Progressive Widening (PW) involves defining when to explore new states in the tree. It is controlled by two parameters $\alpha$ and $k$. The selection criteria is as follows:
\begin{itemize}
\item Compute $k' = k N(h,a)^\alpha$.
\item If $k'$ is greater than the number of children of the node $(h,a)$, then we sample a new state. Otherwise, we choose a state that has already been visited.
\end{itemize}
The branching factor of the search tree can be affected by tuning $k$ and $\alpha$. When the noise in the generative model is large, one typically wants a large branching factor (which can be achieved by increasing $\alpha$, for example). 

%%%%%%%%%%%%%%%%%%%%%%%%%%%%%%%%%%%%%%%%%%%%%%%%%%%%%%%%%%%%%%%%%%%%%%%%%%%%%%%%
\section{Experimental Setup}

\subsection{Simulation and Parameters}

We used the SUMO simulator \cite{sumo} for our experiments, which relies on an Intelligent Driver Model (IDM) \cite{idm-def}. As a consequence, our generative model (relying on Gaussian dynamics) is different than the model used in the test environment. The motivation for this mismatch is to assess how the POMCP algorithm can handle model discrepancies that would exist in real-world applications.  %A block diagram of the simulation loop is represented in \cref{fig:simu}.

%\begin{figure}
%    \centering
%    \resizebox{\columnwidth}{!}{
%    \input{block_diag.tex}
%    }
%    \caption{Simulation loop used to evaluate the policies}
%    \label{fig:simu}
%\end{figure}

The SUMO simulator takes into account the interaction between drivers, and it outputs the position, velocity, and orientation of the vehicles. To simulate the perception of the autonomous car, we added white noise to the position and velocity measurements. The simulation parameters are given in \cref{tab:exp-param}. The traffic density is expressed as the probability of a vehicle going through the intersection every second. The noise parameters correspond to the standard deviations $\sigma_p$ and $\sigma_v$ in \cref{sec:observation-model}.

\begin{table}[h!]
    \centering
    \caption{Parameters of the simulation environment}
    \begin{tabular}{ll}
      \toprule[1pt]
      Parameter   & Value \\
      \midrule
      Traffic density   & \SI{0.2}{} \\
      Position sensor noise  & \SI{0.1}{\meter} \\
      Velocity sensor noise  & \SI{0.1}{\meter\per\second} \\
      Maximum Speed & \SI{13.88}{\meter\per\second} \\
	  \bottomrule[1pt]
	\end{tabular}
    \label{tab:exp-param}
\end{table}

The sequential decision making process proceeds as follows. We start with a prior belief $b_t$. From this belief, we compute an optimal action using the POMCP algorithm and then run a simulation step in SUMO where the environment evolves (including the ego car with respect to the action taken). After this step, the agent receives a reward and observes the environment, and updates its belief to $b_{t+1}$ using the IMM algorithm. The decision and measurement are made every \SI{0.25}{\second}. In order to compute the action at each step, we used the POMCP algorithm with progressive widening with the parameters in \cref{tab:tree}.

\begin{table}[h!]
\centering
\vspace{4pt}
\caption{POMCP solver parameters used in the experiment}
\begin{tabular}{ll}
\toprule[1pt]
Parameter & Value \\
\midrule
Depth & \num{15} \\
Exploration constant & \num{20.0} \\
Tree queries & \num{2000} \\
Rollout policy & TTC Policy \\
PW $\alpha$ & \num{0.2} \\
PW $k$      & \num{4.0} \\
\bottomrule[1pt]
\end{tabular}
\label{tab:tree}
\end{table}

\subsection{Performance Metrics}

To evaluate performance, we used the following metrics:
\begin{itemize}
	\item Average number of collisions % that occurred during the tests (in \%)
	\item Average time to cross the intersection %(in \SI{}{\second})
	\item Success rate at which the car crosses the intersection without crashing or a timeout.
	\item Average time when the traffic is braking %(i.e., any of the car has its brake lights on)
	\item Average time when a car is stopped.
\end{itemize}
The first three metrics account for safety and efficiency. The braking time and the waiting time captures the impact of the ego car on the current traffic.

\subsection{Baseline Policy}

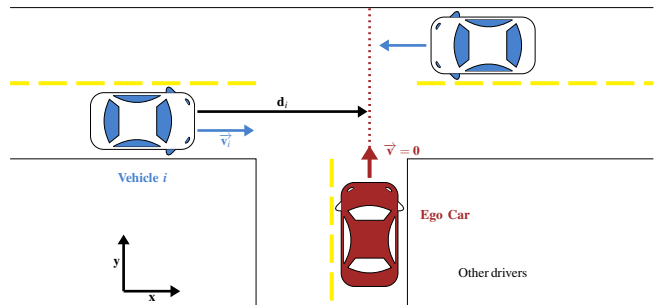
\begin{figure}[h!]
    \centering
    \begin{tikzpicture}

	\definecolor{myblue}{RGB}{72,130,206}
	\definecolor{myred}{RGB}{165,40,40}
	
	%upper border
	\draw [very thick] (-5,16) -- (29,16);
	
	%lower left 
	\draw [very thick] (-5,8) -- (8,8) -- (8,0);
	
	%lower right  
	\draw [very thick] (16,0) -- (16,8) -- (29,8);
	
	%yellow lines 
	\draw [line width=6, color=yellow] (-5,12) -- (-4,12);
	\draw [line width=6, color=yellow] (-3.5,12) -- (-2,12);
	\draw [line width=6, color=yellow] (-1.5,12) -- (0,12);
	\draw [line width=6, color=yellow] (0.5,12) -- (2,12);
	\draw [line width=6, color=yellow] (2.5,12) -- (4,12);
	\draw [line width=6, color=yellow] (4.5,12) -- (6,12);
	\draw [line width=6, color=yellow] (6.5,12) -- (8,12);
	
	\draw [line width=6, color=yellow] (16.5,12) -- (18,12);
	\draw [line width=6, color=yellow] (18.5,12) -- (20,12);
	\draw [line width=6, color=yellow] (20.5,12) -- (22,12);
	\draw [line width=6, color=yellow] (22.5,12) -- (24,12);
	\draw [line width=6, color=yellow] (24.5,12) -- (26,12);
	\draw [line width=6, color=yellow] (26.5,12) -- (28,12);
	\draw [line width=6, color=yellow] (28.5,12) -- (29,12);	
	
	\draw [line width=6, color=yellow] (12,0.5) -- (12,2);
	\draw [line width=6, color=yellow] (12,2.5) -- (12,4);
	\draw [line width=6, color=yellow] (12,4.5) -- (12,6);
	\draw [line width=6, color=yellow] (12,6.5) -- (12,8);

	%insert ego car 
	\node [inner sep=0pt] (egoCar) at (14,4)
	{\includestandalone[]{red_car}};
	
	%insert human cars 
	\node [inner sep=0pt] (hcar2) at (2,10)
	{\includestandalone[angle=-90]{blue_car}}; 
	
	\node [inner sep=0pt] (hcar3) at (20,14)
	{\includestandalone[angle=90]{blue_car}};
	
	%imaginary line
	\draw [line width=4, color=myred, rounded corners=60, loosely dashed] (14,7) -- (14,16);
	
	% velocity vector v
	\draw [line width=6, color=myred, ->, >=Triangle,-{>[scale=1.0]}] (14,7) -- (14,8.7);
	\node (v) [color=myred] at (15.7,8.5) {\huge $\mathbf{\overrightarrow{\mathbf{v}}=0}$};
	
	%distance di
	\draw [line width=4, <->, >=Triangle,-{>[scale=1.0]}] (4.9,10.5) -- node (d) [above] {\huge $\mathbf{d}_i$}(13.9,10.5);
	
	%velocity vi
	\draw [line width=4,color=myblue, ->, >=Triangle,-{>[scale=1.0]}] (4.9,9.5) -- node (vi) [below] {\huge $\overrightarrow{\mathbf{v}_i}$}(7.9,9.5);
	
	% velcity
	\draw [line width=4,color=myblue, ->, >=Triangle,-{>[scale=1.0]}] (17.,14) -- (14.5,14);
	
	%labels
	\node (egoLab) [minimum width=1cm,minimum height=0.5cm, color=myred] at (18,5) {\huge \textbf{Ego Car}};

	\node (humLab) [minimum width=1cm,minimum height=0.5cm,color=myblue] at (2,7) {\huge \textbf{Vehicle \textit{i}}};
	\node (humLab2) [text width=7cm] at (22.2,2) {\huge Other drivers};
	
	% axis 
	\draw [line width=4, ->,>=Triangle,-{>[scale=1.0]}] (1.,1.) -- node(y)[left]{\huge $\mathbf{y}$} (1.,4.);
	\draw [line width=4, ->,>=Triangle,-{>[scale=1.0]}] (1.,1.) -- node(x)[below]{\huge $\mathbf{x}$} (4.,1.);

\end{tikzpicture}
    \caption{Representation of the first phase of the TTC policy, the ego car measures $V_i$ and $d_i$ to compute the TTC and then decides to cross or not according to the threshold.}
    \label{fig:ttc}
\end{figure}

We defined a simple heuristic policy to serve as a baseline that uses a time to collision (TTC) threshold to decide when to cross. The TTC is defined as follows. Consider an imaginary line starting from the ego car aligned with the $y$ axis. The TTC with another vehicle $i$ in the intersection will be the time it takes for the vehicle to reach that line.
For vehicle $i$, it is estimated by dividing $d_i$ by $V_i$, where $d_i$ is the distance indicated in \cref{fig:ttc} and $V_i$ is the speed of vehicle $i$ relative to the ego car.
If the TTC exceeds a threshold for two consecutive time steps of \SI{0.1}{\second}, the vehicle starts the crossing phase. The crossing phase follows the IDM. %The threshold used was \SI{4.5}{\second} in order to guarantee success for both the right and the left turn under the experimental traffic condition in \cref{tab:exp-param}.

%%%%%%%%%%%%%%%%%%%%%%%%%%%%%%%%%%%%%%%%%%%%%%%%%%%%%%%%%%%%%%%%%%%%%%%%%%%%%%%%

\section{Results}

%\subsection{Baseline Analysis}

\begin{figure}
	\centering
	\resizebox{\columnwidth}{!}{
	\begin{tikzpicture}[font=\footnotesize]
\begin{groupplot}[group style={horizontal sep = 1cm, vertical sep = 1.5cm, group size=2 by 2},width=0.5\columnwidth]
\nextgroupplot [grid, xlabel = {TTC threshold (\si{\second})}, ylabel = {Collision rate (\si{\percent})}]\addplot+ [thick, black, mark=none, mark size = {1}, mark options={fill=black}]coordinates {
(0.0, 20.200000000000003)
(0.1, 20.1)
(0.2, 19.900000000000002)
(0.3, 19.6)
(0.4, 19.8)
(0.5, 21.8)
(0.6, 21.0)
(0.7, 21.9)
(0.8, 19.5)
(0.9, 20.9)
(1.0, 22.0)
(1.1, 20.0)
(1.2, 19.400000000000002)
(1.3, 18.099999999999998)
(1.4, 21.0)
(1.5, 17.9)
(1.6, 19.1)
(1.7, 17.2)
(1.8, 16.3)
(1.9, 16.3)
(2.0, 17.2)
(2.1, 16.900000000000002)
(2.2, 12.9)
(2.3, 13.100000000000001)
(2.4, 12.5)
(2.5, 9.5)
(2.6, 8.5)
(2.7, 6.1)
(2.8, 5.4)
(2.9, 3.6999999999999997)
(3.0, 4.0)
(3.1, 2.9000000000000004)
(3.2, 1.5)
(3.3, 1.0)
(3.4, 1.5)
(3.5, 0.8999999999999999)
(3.6, 0.4)
(3.7, 0.4)
(3.8, 0.1)
(3.9, 0.1)
(4.0, 0.0)
(4.1, 0.3)
(4.2, 0.1)
(4.3, 0.1)
(4.4, 0.1)
(4.5, 0.0)
(4.6, 0.0)
(4.7, 0.1)
(4.8, 0.0)
(4.9, 0.0)
(5.0, 0.0)
};
\nextgroupplot [grid, xlabel = {TTC threshold (\si{\second})}, ylabel = {Average time to cross (\si{\second})}]\addplot+ [thick, black, mark=none, mark size = {1}, mark options={fill=black}]coordinates {
(0.0, 7.372199999999999)
(0.1, 7.439900000000001)
(0.2, 7.462600000000001)
(0.3, 7.507200000000001)
(0.4, 7.5188999999999995)
(0.5, 7.467100000000001)
(0.6, 7.555500000000001)
(0.7, 7.515600000000001)
(0.8, 7.6806)
(0.9, 7.703100000000001)
(1.0, 7.6478)
(1.1, 7.789700000000001)
(1.2, 7.896800000000001)
(1.3, 8.0375)
(1.4, 7.9261)
(1.5, 8.0919)
(1.6, 8.176300000000001)
(1.7, 8.3181)
(1.8, 8.3216)
(1.9, 8.4323)
(2.0, 8.5)
(2.1, 8.607899999999999)
(2.2, 8.9469)
(2.3, 8.9324)
(2.4, 9.093300000000001)
(2.5, 9.3421)
(2.6, 9.4818)
(2.7, 9.6904)
(2.8, 9.752500000000001)
(2.9, 9.962200000000001)
(3.0, 10.0327)
(3.1, 10.1788)
(3.2, 10.3332)
(3.3, 10.404900000000001)
(3.4, 10.807400000000001)
(3.5, 10.6835)
(3.6, 10.6481)
(3.7, 10.7943)
(3.8, 11.0356)
(3.9, 11.297400000000001)
(4.0, 11.3499)
(4.1, 11.298300000000001)
(4.2, 11.649500000000002)
(4.3, 11.3109)
(4.4, 11.8801)
(4.5, 11.950800000000001)
(4.6, 11.5678)
(4.7, 11.935500000000001)
(4.8, 12.152500000000002)
(4.9, 11.9479)
(5.0, 12.3247)
};
\nextgroupplot [grid, xlabel = {TTC threshold (\si{\second})}, ylabel = {Average braking time (\si{\second})}]\addplot+ [thick, black, mark=none, mark size = {1}, mark options={fill=black}]coordinates {
(0.0, 0.767)
(0.1, 0.8246000000000001)
(0.2, 0.7676000000000001)
(0.3, 0.8132)
(0.4, 0.7486)
(0.5, 0.8006000000000001)
(0.6, 0.7969)
(0.7, 0.7922)
(0.8, 0.7693)
(0.9, 0.8081)
(1.0, 0.7496)
(1.1, 0.7504)
(1.2, 0.7719)
(1.3, 0.7259000000000001)
(1.4, 0.7009000000000001)
(1.5, 0.6983)
(1.6, 0.7266)
(1.7, 0.6547000000000001)
(1.8, 0.6163000000000001)
(1.9, 0.5811000000000001)
(2.0, 0.6414)
(2.1, 0.6605000000000001)
(2.2, 0.6489)
(2.3, 0.6136)
(2.4, 0.5854)
(2.5, 0.6097000000000001)
(2.6, 0.6005)
(2.7, 0.4791000000000001)
(2.8, 0.5082)
(2.9, 0.49970000000000003)
(3.0, 0.46820000000000006)
(3.1, 0.4215)
(3.2, 0.3996)
(3.3, 0.35150000000000003)
(3.4, 0.3831)
(3.5, 0.33220000000000005)
(3.6, 0.3371)
(3.7, 0.3128)
(3.8, 0.2585)
(3.9, 0.2215)
(4.0, 0.2331)
(4.1, 0.2129)
(4.2, 0.2041)
(4.3, 0.18720000000000003)
(4.4, 0.17300000000000001)
(4.5, 0.14170000000000002)
(4.6, 0.1335)
(4.7, 0.1323)
(4.8, 0.1004)
(4.9, 0.0825)
(5.0, 0.0974)
};
\nextgroupplot [grid, xlabel = {TTC threshold (\si{\second})}, ylabel = {Average waiting time (\si{\second})}]\addplot+ [thick, black, mark=none, mark size = {1}, mark options={fill=black}]coordinates {
(0.1, 0.0013)
(0.2, 0.0015)
(0.3, 0.0033000000000000004)
(0.4, 0.0026)
(0.5, 0.004200000000000001)
(0.6, 0.0037)
(0.7, 0.0017000000000000001)
(0.8, 0.001)
(0.9, 0.0019)
(1.0, 0.0016)
(1.1, 0.003)
(1.2, 0.002)
(1.3, 0.0014000000000000002)
(1.4, 0.0009)
(1.5, 0.0015)
(1.6, 0.0017000000000000001)
(1.7, 0.0026)
(1.8, 0.0008)
(1.9, 0.0017000000000000001)
(2.0, 0.0017000000000000001)
(2.1, 0.0017000000000000001)
(2.2, 0.0006000000000000001)
(2.3, 0.0014000000000000002)
(2.4, 0.0007000000000000001)
(2.5, 0.0006000000000000001)
(2.6, 0.0011)
(2.7, 0.0006000000000000001)
(2.8, 0.0013)
(2.9, 0.001)
(3.0, 0.0007000000000000001)
(3.1, 0.0008)
(3.2, 0.0007000000000000001)
(3.3, 0.0013)
(3.4, 0.0005)
(3.5, 0.0007000000000000001)
(3.6, 0.0012000000000000001)
(3.7, 0.0016)
(3.8, 0.001)
(3.9, 0.0014000000000000002)
(4.0, 0.0011)
(4.1, 0.0009)
(4.2, 0.0021000000000000003)
(4.3, 0.0013)
(4.4, 0.0014000000000000002)
(4.5, 0.0007000000000000001)
(4.6, 0.001)
(4.7, 0.0016)
(4.8, 0.0013)
(4.9, 0.0011)
(5.0, 0.0013)
};
\end{groupplot}

\end{tikzpicture}
	}
	\caption{Four metrics with different TTC thresholds for right-turns with a traffic density of \num{0.2}.}
	\label{fig:threshold}
\end{figure}
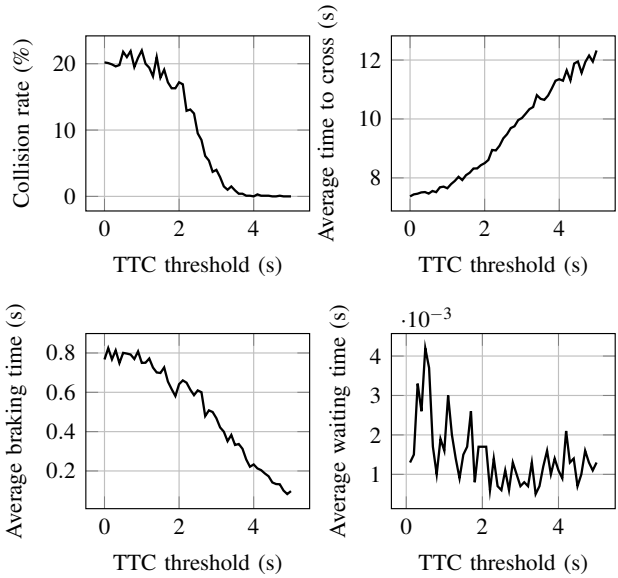

We analyzed the influence of the TTC threshold on different metrics and chose the threshold that results in zero collisions over a thousand simulations. \Cref{fig:threshold} shows how four metrics vary with respect to the choice of the TTC threshold. Due to measurement noise, we can see that even for high thresholds there are still small fluctuations in the collision rate. The chosen threshold for comparison with the POMCP policy is \SI{4.5}{\second} in order to guarantee success for both the right and the left turns under the experimental traffic condition in \cref{tab:exp-param}, without being overly conservative.

%\subsection{Behavior Tuning}

\begin{figure}[h]
	\begin{tikzpicture}[font=\footnotesize]
\begin{axis}[legend pos=north east,grid,width = 0.8\columnwidth, xlabel = {Average time to cross (\si{\second})}, ylabel = {Collision rate (\si{\percent})}]
\addplot+ [blue, dashed, thick, mark={none},mark options={fill=gray}, mark size = {1}]coordinates {
(7.372199999999999, 20.200000000000003)
(7.439900000000001, 20.1)
(7.462600000000001, 19.900000000000002)
(7.467100000000001, 21.8)
(7.507200000000001, 19.6)
(7.515600000000001, 21.9)
(7.5188999999999995, 19.8)
(7.555500000000001, 21.0)
(7.6478, 22.0)
(7.6806, 19.5)
(7.703100000000001, 20.9)
(7.789700000000001, 20.0)
(7.896800000000001, 19.400000000000002)
(7.9261, 21.0)
(8.0375, 18.099999999999998)
(8.0919, 17.9)
(8.176300000000001, 19.1)
(8.3181, 17.2)
(8.3216, 16.3)
(8.4323, 16.3)
(8.5, 17.2)
(8.607899999999999, 16.900000000000002)
(8.9324, 13.100000000000001)
(8.9469, 12.9)
(9.093300000000001, 12.5)
(9.3421, 9.5)
(9.4818, 8.5)
(9.6904, 6.1)
(9.752500000000001, 5.4)
(9.962200000000001, 3.6999999999999997)
(10.0327, 4.0)
(10.1788, 2.9000000000000004)
(10.3332, 1.5)
(10.404900000000001, 1.0)
(10.6481, 0.4)
(10.6835, 0.8999999999999999)
(10.7943, 0.4)
(10.807400000000001, 1.5)
(11.0356, 0.1)
(11.297400000000001, 0.1)
(11.298300000000001, 0.3)
(11.3109, 0.1)
(11.3499, 0.0)
(11.5678, 0.0)
(11.649500000000002, 0.1)
(11.8801, 0.1)
(11.935500000000001, 0.1)
(11.9479, 0.0)
(11.950800000000001, 0.0)
(12.152500000000002, 0.0)
(12.3247, 0.0)
};
\addlegendentry{TTC policy}
\addplot+ [black, thick, mark ={none}, mark options={fill=black}, mark size = {1}]coordinates {
(7.19825, 25.4)
(8.18375, 9.0)
%(8.19375, 12.5)
(8.3285, 0.1)
(8.4515, 0.0)
(8.46825, 0.0)
(8.58125, 5.5)
(8.63875, 3.0)
(8.744, 4.3999999999999995)
(8.80875, 3.0)
(8.848, 1.9)
(8.87625, 0.1)
(8.90475, 0.0)
(8.92025, 0.0)
(8.963, 0.3)
(8.98825, 0.2)
(9.003, 0.0)
(9.01425, 0.2)
(9.06075, 0.3)
(10.68875, 0.0)
(21.3405, 0.0)
};
\addlegendentry{POMCP policy}
\end{axis}

\end{tikzpicture}
	\centering 
	\caption{Trade-off between collision rate and time to cross as key parameters are varied for the policies (action cost for POMCP policy and threshold for TTC policy) in the right-turn scenario with a traffic density of \num{0.2}.}
	\label{fig:tradeoff}
\end{figure}
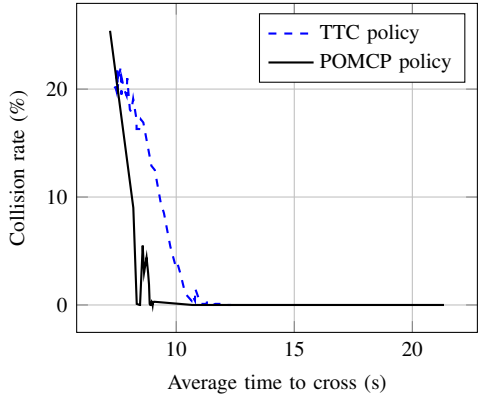

The reward function in the POMDP formulation can be used to tune the behavior of the agent in favor one objective over another. By varying the reward parameters, such as the cost for each action, we can reach a point where the expected reward will decrease very fast with time. As a consequence, increasing this cost will favor a minimization of the time to cross at the expense of collision risk.  By tuning the reward function, we can balance these two conflicting objectives.
% Display the evolution of the metrics with the action cost
% Display the trade-off graph.
\Cref{fig:tradeoff} illustrates the trade-off between these two objectives for the right-turn scenario. The POMCP policy clearly dominates the threshold policy with respect to the collision rate and the time to cross. 

%\subsection{Comparison of the POMCP policy with the baseline}

We selected a conservative set of POMCP policy parameters (\cref{tab:reward}) to compare against the TTC policy with a threshold of \SI{4.5}{\second}. The penalties chosen for each action differ in order to favor forward motion. The numerical values were chosen to separate each outcome by several orders of magnitude. We also compared against a random policy for both scenarios in \cref{tab:right-results} and \cref{tab:left-results} for fixed traffic conditions (\cref{tab:exp-param}). The metrics are averaged over one thousand simulations.

For the right-turn scenario, both the POMCP and TTC policies achieve 100\% success rate (\cref{tab:right-results}). However, the POMCP policy outperforms the TTC policy in average time to cross the intersection. The waiting time is higher for the POMCP policy, but it still does not exceed \SI{10}{\milli\second} on average. 

\Cref{tab:left-results} shows that for the left-turn scenario the TTC policy achieved zero collisions and 100\% success rate, whereas the POMCP policy still has some collisions (\SI{0.2}{\%}) and time-outs, leading to a success rate of (\SI{99.0}{\%}). The braking time and waiting time are also higher.

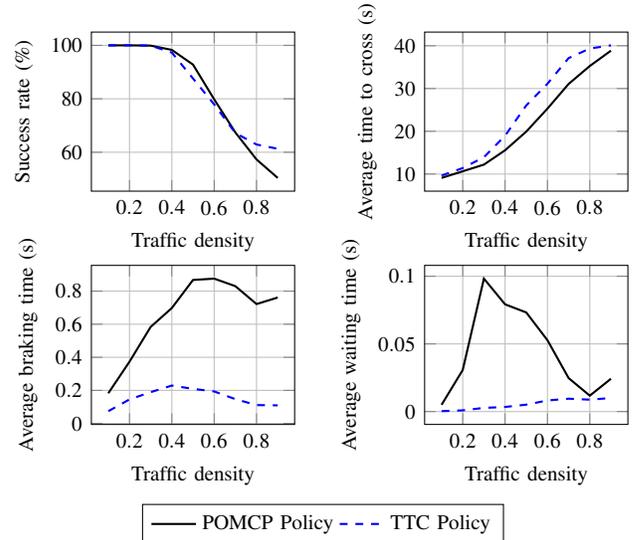
\begin{figure}
	\centering
	\resizebox{\columnwidth}{!}{
	\begin{tikzpicture}[font=\footnotesize]
\begin{groupplot}[grid, group style={horizontal sep = 1.75cm, vertical sep = 1cm, group size=2 by 2},width=0.5\columnwidth]
\nextgroupplot [xlabel = {Traffic density}, ylabel = {Success rate (\si{\percent})}, legend style={at={(0.02,0.02)},anchor=south west}]
\addplot+ [mark = {none}, thick, black, mark size = {1}]coordinates {
(0.1, 100.0)
(0.2, 100.0)
(0.3, 99.9)
(0.4, 98.3)
(0.5, 92.8)
(0.6, 79.8)
(0.7, 67.4)
(0.8, 57.300000000000004)
(0.9, 50.4)
};
%\addlegendentry{POMCP Policy}
\addplot+ [mark = {none}, dashed, thick, blue, mark size = {1}, mark options={fill=gray}]coordinates {
(0.1, 100.0)
(0.2, 100.0)
(0.3, 99.9)
(0.4, 97.3)
(0.5, 87.6)
(0.6, 77.9)
(0.7, 67.19999999999999)
(0.8, 62.9)
(0.9, 61.3)
};
%\addlegendentry{TTC Policy}
\nextgroupplot [grid, xlabel = {Traffic density}, ylabel = {Average time to cross (\si{\second})}]
\addplot+ [mark = none, thick, black, mark size = {1}, mark options={fill=gray}]coordinates {
(0.1, 9.0985)
(0.2, 10.6465)
(0.3, 12.2055)
(0.4, 15.5605)
(0.5, 19.9275)
(0.6, 25.29225)
(0.7, 31.06725)
(0.8, 35.264)
(0.9, 38.855)
};
%\addlegendentry{POMCP Policy}
\addplot+ [mark = none, dashed, thick, blue, mark size = {1}, mark options={fill=gray}]coordinates {
(0.1, 9.6492)
(0.2, 11.402500000000002)
(0.3, 13.9398)
(0.4, 18.9564)
(0.5, 26.048200000000005)
(0.6, 31.110000000000003)
(0.7, 37.111399999999996)
(0.8, 39.34780000000001)
(0.9, 40.093700000000005)
};
%\addlegendentry{TTC Policy}
\nextgroupplot [grid, xlabel = {Traffic density}, ylabel = {Average braking time (\si{\second})}]
\addplot+ [mark = none, thick, black, mark size = {1}, mark options={fill=gray}]coordinates {
(0.1, 0.18275)
(0.2, 0.37325)
(0.3, 0.5835)
(0.4, 0.69775)
(0.5, 0.8675)
(0.6, 0.8755)
(0.7, 0.83)
(0.8, 0.72175)
(0.9, 0.76125)
};
%\addlegendentry{POMCP Policy}
\addplot+ [mark = none, dashed, thick, blue, mark size = {1},mark options={fill=gray}]coordinates {
(0.1, 0.0741)
(0.2, 0.1452)
(0.3, 0.18910000000000002)
(0.4, 0.2291)
(0.5, 0.20910000000000004)
(0.6, 0.1935)
(0.7, 0.1469)
(0.8, 0.11100000000000002)
(0.9, 0.10930000000000001)
};
%\addlegendentry{TTC Policy}
\nextgroupplot [grid, xlabel = {Traffic density}, ylabel = {Average waiting time (\si{\second})}, tick label style={/pgf/number format/fixed}, legend style={at={(-0.5,-0.5)},
	anchor=north,legend columns=-1}]
\addplot+ [mark = none, thick, black, mark size = {1}]coordinates {
(0.1, 0.005)
(0.2, 0.03075)
(0.3, 0.09825)
(0.4, 0.07925)
(0.5, 0.07325)
(0.6, 0.05275)
(0.7, 0.02475)
(0.8, 0.01175)
(0.9, 0.02425)
};
\addlegendentry{POMCP Policy}
\addplot+ [mark = none, dashed, thick, blue, mark size = {1}, mark options={fill=gray}]coordinates {
(0.1, 0.00030000000000000003)
(0.2, 0.0009)
(0.3, 0.0027)
(0.4, 0.0034000000000000002)
(0.5, 0.0051)
(0.6, 0.0082)
(0.7, 0.009500000000000001)
(0.8, 0.0088)
(0.9, 0.010000000000000002)
};
\addlegendentry{TTC Policy}
\end{groupplot}

\end{tikzpicture}
	}
	\caption{Metrics for varying traffic density for both the TTC and the POMCP policies in the right-turn scenario.}
	\label{fig:density}
\end{figure}

In order to assess the scalability of the two policies, we analyzed the evolution of the metrics with an increasing traffic density for the right-turn scenario. Both policies achieved a collision rate of \num{0}\%, but they were subject to time-outs, which are reflected by a decrease in the success rate as the traffic density increases. \Cref{fig:density} shows that until a density of \num{0.7}, the POMCP policy has a success rate at least as high as the TTC policy. For every tested traffic density, the POMCP policy takes on average less time to cross the intersection with a maximum difference of \SI{6.12}{\second} for a traffic density of \num{0.5}. However, the braking time and the waiting time are higher than the TTC policy.

\begin{table}[h]
	\centering
	\caption{Reward function parameters}
	\label{tab:reward}
	\begin{tabular}{lS}
		\toprule[1pt]
		Parameter & \text{Value} \\
		\midrule
		Collision penalty & -2000.0 \\
		Acceleration penalty   & -4.98 \\
		Maintaining speed penalty & -4.99\\
		Moderate braking penalty & -5.0 \\
		Strong braking penalty & -5.02 \\
		Crossing reward & +100.0 \\
		\bottomrule[1pt] 
    \end{tabular}
	
\end{table}

\begin{table}[h]
    \centering
    \vspace{4pt}
    \caption{Performance of the policies for the right-turn scenario}
    \label{tab:right-results}
    \resizebox{\columnwidth}{!}{
    \begin{tabular}{lSSSSS}% *{4}{S[table-format=2.4]}}
    \toprule[1pt]
     Policy    & \text{Time} & \text{Braking} & \text{Waiting} & \text{Collision} & \text{Success}\\
       & \text{to cross (s)} & \text{time (s)}  & \text{time (s)}  & \text{rate (\%)} & \text{rate (\%)}\\
     \midrule
     POMCP     & 10.6465  & 0.3733 & 0.0308 & 0.0 & 100.0\\
     TTC  & 10.7270 & 0.1170 & 0.0014 & 0.0 & 100.0\\
     Random   &  55.5948  & 37.9805 & 20.4013 & 9.80 & 0.40 \\
     \bottomrule[1pt]
    \end{tabular}
    }
        
\end{table}

\begin{table}[h]
    \centering
    \caption{Performance of the policies for the left-turn scenario}
    \label{tab:left-results}
    \resizebox{\columnwidth}{!}{
    \begin{tabular}{lSSSSS}
    \toprule[1pt]
       Policy    & \text{Time} & \text{Braking} & \text{Waiting} & \text{Collision} & \text{Success}\\
      & \text{to cross (s)} & \text{time (s)}  & \text{time (s)}  & \text{rate (\%)} & \text{rate (\%)}\\
      \midrule
     POMCP     & 10.3735 & 0.3763 & 0.1293 & 0.2 & 99.0\\
     TTC  & 10.7704 & 0.1912 & 0.0010 & \bfseries 0.0 & 100.0\\
     Random   & 34.1028 &  17.0503 & 10.2940 & 75.70 & 0.10\\
     \bottomrule[1pt]
    \end{tabular}
    }    
\end{table}

%\subsection{Discussion}

The results show that both the POMCP policy and the TTC policy are safe for the right-turn scenario, even under some measurement noise. Moreover, the POMCP policy reaches the goal faster than the TTC policy but will cause the other users to brake and wait more often. \Cref{fig:density} shows that the POMCP policy manages to cross the intersection more often and faster than the TTC policy up to a certain traffic density at the expense of somewhat greater disruption to the traffic. 
The choice of the reward function penalizing the actions makes the ego car more eager to enter the intersection and is not penalizing for making the other drivers brake or wait. The problem formulation can explain the higher braking time and waiting time for the POMCP policy.

For left turns, similar conclusions can be drawn on the time to cross, the braking time, and the waiting time. However, the POMCP policy still has some collisions. One  explanation is the difference between the generative model and the simulator model. In order to assess the discrepancies between the two, we measured the error in the position prediction as a function of the planning horizon. We found an average error of \SI{2.15}{\meter} when predicting the motion of the other cars ten steps ahead (\SI{2.5}{\second}). %\Cref{fig:error} shows the growth of the error as the planning depth increases. 
This difference in the predicted position prevents the algorithm from predicting some rapid maneuver changes. A more sophisticated generative model, combined with a good internal state estimator could improve performance, and we believe zero collision rate could be achieved for left turns.

%\begin{figure}
%	\centering
%	\input{error.tex}
%	\caption{Average error in position when propagating the transition model for predicting the future position of other vehicles. The error is averaged over 50 cars during \SI{250}{\second} of simulation in SUMO.}
%	\label{fig:error}
%\end{figure}

%%%%%%%%%%%%%%%%%%%%%%%%%%%%%%%%%%%%%%%%%%%%%%%%%%%%%%%%%%%%%%%%%%%%%%%%%%%%%%%%
\section{Conclusions}

We demonstrated an online belief state planning approach based on a POMDP formulation to address a decision problem at an unsignalized intersection. The proposed approach performs better than a heuristic policy, even with a fairly simple transition model. It is robust enough to handle discrepancies between the assumed model and the simulator model. The resulting policy outperforms the baseline in most of the considered metrics. We showed that we can balance the different objectives of the problem by tuning the reward function of the POMDP. Further work will involve improving the state estimator by using a more accurate generative model than the linear Gaussian model used in this paper. We would also like to complement our current approach with an offline planner and increase the complexity of the scenarios and move toward more realistic models. An immediate consideration would involve pedestrians, as well as sensor occlusions.

\printbibliography

\addtolength{\textheight}{-12cm}   % This command serves to balance the column lengths

\end{document}